%% file: main-10171-Goodale.tex
\documentclass[11pt,a4paper,acceptedWithA]{article}
\usepackage{latexsym}
\usepackage{mathptmx}
\usepackage{url}
\usepackage[T1]{fontenc}

\usepackage[acceptedWithA]{tacl2021v1}
\usepackage{xspace,mfirstuc,tabulary}

\newif\iftaclinstructions
\taclinstructionsfalse % AUTHORS: do NOT set this to true
\iftaclinstructions
\renewcommand{\confidential}{}
\renewcommand{\anonsubtext}{(No author info supplied here, for consistency with
TACL-submission anonymization requirements)}
\newcommand{\instr}
\fi

\iftaclpubformat % this "if" is set by the choice of options

\else

\fi

\author{
   Michael Goodale
  \and
  Salvador Mascarenhas
  \\
  \ \\
  Institut Jean Nicod, Département d’études cognitives
  \\ENS, EHESS, CNRS, PSL University.
  \\
  \texttt{\{michael.goodale, salvador.mascarenhas\}@ens.fr}
}
\newcommand{\parencite}[1]{\citep{#1}}
\newcommand{\textcite}[1]{\citet{#1}}
\date{}

\usepackage{caption}
\usepackage{subcaption}
\usepackage{mathtools}

\usepackage{tikz}
\usetikzlibrary{fit}
\usetikzlibrary{arrows}
\usetikzlibrary{calc}
\usetikzlibrary{positioning}
\usepackage{amsmath}
\usepackage{enumitem}
\usepackage{standalone}

\usepackage{booktabs}
\usepackage{hyphenat}
\usepackage{amssymb}
\usepackage{stmaryrd}
\usepackage{xcolor}
\definecolor{green}{RGB}{34,139,34}
\DeclarePairedDelimiter{\set}{\{}{\}}

\renewcommand{\phi}{\varphi}
\newcommand{\mybar}{\ensuremath{\vert}}

\title{Fodor and Pylyshyn's Systematicity Challenge Still Stands}
\begin{document}
\maketitle
\begin{abstract}
  The recent successes of neural networks producing human-like language have caused significant stir in cognitive science, with many researchers arguing that classical puzzles about human cognition and challenges to artificial intelligence are being solved by neural networks.
  A notable case is the argument from systematicity due to Jerry Fodor and Zenon Pylyshyn, argues that humans display systematic biconditional dependencies.
  For example, someone can understand the sentence ``John saw Mary'' just in case that they understand the sentence ``Mary saw John.''
  Symbolic systems explain this systematicity of language and thought, while neural networks offer no immediate explanation.
  Several recent articles argue that this challenge has now been met by neural networks.
  In particular, Brenden Lake and Marco Baroni argue that their \emph{meta-learning for compositionality} protocol matches and perhaps explains human systematicity.
  We demonstrate that these conclusions are premature.
  Among other results, we found that their model struggles to learn rules that are even slightly out of distribution compared to their training data.
  Furthermore, the model behaves unsystematically even on many within-distribution problems.
  We conclude that Fodor and Pylyshyn's challenge to neural networks remains unmet.
\end{abstract}

\section{Introduction}

The nineteen-eighties witnessed the maturation of a model of computation that revolutionized cognitive science and machine learning, and arguably a few decades later society at large: connectionism \parencite{Rumelhart:1986}.
Unlike classical symbolic models, connectionist models looked mappable to neurobiological structures, while being, in principle, capable of performing any computation classical models could.
They showed great potential to reconstruct most or all of the key insights of symbolic models \parencite{Smolensky:1988}.

Three decades later, connectionist models of one form or another constitute the computational basis for the large language models that have so strongly captured our attention, both in the ivory tower and in society at large.
In particular, some cognitive scientists have argued that these architectures currently provide our best theories of natural language \parencite{Piantadosi:2023}, a human capacity traditionally thought to require symbolic models as a matter of necessity.
% More recently, models in this class have captured public attention, and many experts and commentators consider that they hold the promise of one day, soon enough, either solving all of society's ills or precipitating our collective demise.

\subsection{The argument from systematicity}
Yet, already in the decade of their invention, connectionist architectures had their detractors.
In a classical article from 1988, Jerry Fodor and Zenon Pylyshyn argued that this class of cognitive theories had a major and \emph{damning} handicap compared to classical symbolic architectures \parencite{Fodor:1988}.
The gist of the argument from \emph{systematicity} is that many distinctively human faculties display biconditional interdependences: a human has faculty $X$ precisely to the extent that they have faculty $Y$.
Natural language offers a crisp illustration:
a human being has the ability to understand the sentence ``Ann introduced Bill to Claire'' precisely to the extent that that same human being can understand the sentences ``Bill introduced Claire to Ann,'' ``Claire introduced Dan to Ed,'' and so forth.
Crucially, this kind of systematicity is no accident, but rather a fundamental feature of cognition that requires explanation.
In particular, there aren't and there cannot be \emph{punctate minds}, for example minds that understand ``Ann introduced Bill to Claire'' but are stumped by ``Bill introduced Claire to Ann.''

The problem, according to Fodor and Pylyshyn, is that connectionist models fail to predict, let alone explain systematicity of this kind, while classical symbolic systems predict and explain it by default.

In a classical symbolic system, the content of ``Ann introduced Bill to Claire'' will be represented by means of a triadic relation $I$ which connects three terms $a$, $b$, and $c$, thus: $I(a, b, c)$.
This representation of the complete proposition (the content of a natural language sentence or a thought) vitally \emph{contains} the relation $I$, as well as each of the terms of the relation, in an explicit structural configuration which can be depicted by a tree.
Thus, a symbolic system that can represent $I(a, b, c)$ can do so only because it can represent the constitutive elements of that symbolic formula in a particular structure.
From this, it follows that by default such a system must be able to represent $I(b, a, c)$ and all other well-formed permutations of the symbols.\footnote{Why the caveat ``by default''?
  Because a symbolic system can be contrived, of course, which blocks one or more of the possible permutations.
  For example, we could partition the set of all terms $T = \set{a, b, c}$ into $A = \set{a}$ and $X = \set{b, c}$, and change the relevant recursive clause from ``if $t, t', t'' \in T$, then $I(t, t', t'')$ is well formed'' to ``if $t \in A$ and $t', t'' \in X$, then $I(t, t', t'')$ is well formed.''
  Clearly though, this definition would be strictly more complex, it would in no way come ``for free,'' requiring substantive motivation of one form or another.
  This is the sense in which classical symbolic systems are systematic by default: symbolic systems can be punctate, but they must go to significant and glaring lengths to do so.
  Dissatisfaction with this \emph{argument from default} led \textcite{calvo_category_2014} to propose an upgrade of the symbolic view using the formalism of category theory to pin down precisely the class of symbolic theories that truly \emph{must} be systematic in the relevant sense.}
Additionally, a symbolic system that represents $I(a, b, c)$ will license all manner of inferences from such a representation, for example $\exists x . I(x, b, c)$ (``Someone introduced Bill to Claire'').
These inferences are crucially licensed by the form of the symbolic representation itself.
That is, the process that takes $I(a, b, c)$ and produces $\exists x . I(x, b, c)$ manipulates the constituents of the original representation, substituting $x$ for $a$, and adding the prefix $\exists x$.
Because inferences in a symbolic architecture operate on the constituents of representations, inference in these systems is, by default, systematic.

By contrast, in a connectionist architecture the distributed network that represents ``Ann introduced Bill to Claire'' does not guarantee by default the possibility of representing ``Bill introduced Ann to Claire.''
This is because a connectionist system does not by default require that ``Ann,'' ``Bill,'' ``Claire,'' and ``introduced'' be represented as such in order to represent ``Ann introduced Bill to Claire.''
Similarly, the inferences that a particular network may license on the basis of the representation for ``Ann introduced Bill to Claire,'' such as ``Someone introduced Bill to Claire,'' do not by default generalize, say to ``Ann introduced someone to Claire.''
Since connectionist systems do not by default represent the (symbolic) constituents of (symbolically) complex representations, the processes of inference that can be defined over these systems do not, by default, involve manipulations of the constituents of complex representations.

There are excellent reasons to be critical of Fodor and Pylyshyn's broader arguments and their conclusion that connectionist systems can at best be a model of how a classical architecture is implemented in the brain.
But the fragment of the argument we present here was accepted as valid and relevant by much of the field, including some of the authors' chief opponents.
In particular \textcite{Smolensky:1988,Smolensky:1987} rejects the bleak conclusion, but wholeheartedly accepts that the systematicity that comes out essentially ``for free'' in symbolic architectures requires considerable ingenuity to bring out in a connectionist architecture.
He proposes that a special class of connectionist systems can adequately describe and explain systematicity, in particular by means of binding units which are tensor products of role vectors and filler vectors, providing an analog of the classical notion of constituent structure.

\subsection{Systematicity, generalization, and behavior}
This much will suffice in way of an exposition of the argument and the debate on systematicity.
Yet, there are a number of concepts in the vicinity of the notion of systematicity that is of interest to us here.
For the purpose of this article then, it is just as important to clarify what the systematicity questions are \emph{not at all} about.
The systematicity debate was \emph{not} about whether it is possible for a neural network to \emph{behave} in a systematic fashion.
Indeed, \textcite{Fodor:1988} are very clear in their main recommendation for connectionists: drop your ambitions of providing a theory at the \emph{cognitive} level, and instead see your work as a theory of the \emph{implementation} of a cognitive-level classical symbolic theory.
Of course, for Fodor and Pylyshyn, humans behave systematically, if any organism does.
If the connectionist architecture is promising as a theory of implementation of a classical symbolic cognitive theory, it follows that neural networks can, for Fodor and Pylyshyn, implement systematic behavior.
Their question was whether they can properly \emph{explain} the systematic properties of cognition, in particular by explaining what is ``wrong'' with punctate minds and why they do not exist.

A simple thought experiment can help clarify the notions of interest here.
Say we have a collection of minds that speak a language that allows for exactly four sentences, which under a symbolic architecture we would represent as $aRa$, $aRb$, $bRa$, and $bRb$.
A connectionist architecture is under no obligation to analyze these sentences as involving two terms $a, b$ and a relation $R$.
Indeed, the simplest way to represent this language is likely with a lookup table: we might as well simply list the four expressions of this language.
Here is Fodor and Pylyshyn's point: a list of these four sentences, absent a symbolic analysis of each of those sentences that accounts for their constituent structure, offers no explanation of \emph{why} all of the minds that speak this language have all four of those sentences, no more and no fewer.
A list implementation of this language, in and of itself, provides no reason for its own completeness, its own systematicity.
It is only through a symbolic analysis of the sentences that a theory can ``see'' the pattern for what it is and explain the systematicity.

Importantly, the question of how this system is or ought to be implemented is unaffected by these considerations: the system might very well be implemented precisely as a list of four holistic expressions.
In fact this is probably the most efficient implementation possible for such a small language.
Nevertheless, if it turns out that \emph{all} of the relevant minds have \emph{all} four of these sentences, and that that completeness or systematicity tells us something important about the nature of these minds, then we need some level of analysis that can formulate this constraint and explain it.
This is what symbolic systems have to offer beyond connectionist systems.

A notion intuitively related to systematicity and much discussed in the machine\hyp{}learning literature is \emph{generalization}, in particular for the special case of linguistic systems of interest to us here \parencite{Kirk:2023,McCoy:2019,Kim:2020,Dubois:2020}.
In a recent broad review of generalization in natural-language processing, \textcite{Hupkes:2023} characterize it as ``the ability to successfully transfer representations, knowledge and strategies from past experience to new experiences.''
We broadly agree with this definition, and note that it is clearly a \emph{behavioral} notion: a system achieves good generalization to the extent that it displays some form of ``success'' in novel ``experiences.''
This notion of generalization is of course related to systematicity, in that we expect systematicity to have behavioral reflections of this kind.
Moreover, from the standpoint of language-model evaluation, it is important to use behavioral measures, since behavior, in neural networks as in humans, constitutes the most immediate source of phenomena to explain, as well as the class of phenomena with most direct relevance for practical applications of our language models.
But it is clear that this notion of generalization does not cover systematicity in the sense it had to everyone involved in the 1980s debates.
We will return to this issue in our Discussion section.

\bigskip\noindent
Let us take stock of what we have so far.
Systematicity is the property of displaying deep biconditional dependencies between faculties, where these dependencies play a central role in the very nature of those faculties.
A punctate mind that can represent $aRb$ but not $bRa$ is, from this standpoint, a qualitatively different kind of mind than a systematic one.
Crucially, systematicity is not simply a behavioral notion.
A mind that can represent all four combinations of our running example in this section would be behaviorally systematic, it would have achieved perfect generality.
Yet, whether or not its systematicity is explained will be a question about our analysis of this mind: is there an account that explains \emph{why} these minds always have all four of these representations?
Symbolic theories immediately do the job by default, while connectionist theories do not, or at least not without considerable mathematical ingenuity, to date invariably reflected in strong architectural constraints.

\subsection{A case study}
In an article recently published in \emph{Nature}, Brenden Lake and Marco Baroni argue that they have developed a particular protocol for meta-training neural networks which resolves this challenge from systematicity \parencite{lake-human-like-2023}.
They focus on transformer models trained on a particular sequence\hyp{}to\hyp{}sequence mapping task, where strings of (nonce) words are mapped to strings of colored discs, and some words correspond to a disc of a particular color, while others are instructions for transforming some of the surrounding colored discs.
% Concretely and by way of example, the model might learn that \texttt{fep} means {\color{blue}\(\bullet\)} and \texttt{blicket} is an instruction to display the preceding disc three times, so that \texttt{fep blicket} would be mapped to the interpretation {\color{blue}\(\bullet\bullet\bullet\)}.
After presenting the details of their meta-learning protocol and their results, we will argue that Lake and Baroni's (\citeyear{lake-human-like-2023}) work fails to address, let alone solve the systematicity challenge to neural networks, for three orders of reasons.

First, their work is effectively exclusively concerned with the \emph{behavior} of neural networks.
At no point in their article or the supplementary materials do Lake and Baroni tell us in what way their meta-learning regimen \emph{guarantees} behavioral systematicity or \emph{explains it} when it emerges.
Yet, Lake and Baroni directly and repeatedly position themselves as answering the challenge raised by Fodor and Pylyshyn.

Second, we argue that the proposed meta-learning protocol, which purports to promote systematicity and compositionality, does not deliver on that promise in its current form.
This is because their protocol is extremely sensitive to details of the data seen during meta-learning, leading to poor generalization outside distribution, compared to standard meta-learning methods like model-agnostic meta-learning.

Third and final, neural networks trained with Lake and Baroni's method fail to even \emph{behave} systematically.
We unfold Fodor and Pylyshyn's conception of systematicity into three distinct ``levels'' of behavioral systematicity: (i) narrow within-distribution behavior, (ii) sensitivity to irrelevant details, e.g., arbitrariness of symbols, and (iii) the ability to be equally systematic with different grammars.
We found that Lake and Baroni's model has behavioral failures of systematicity on all three levels, which we will present in detail.
% In sum, we argue that Lake and Baroni's \emph{meta-learning for compositionality} in no way constitutes progress towards successfully addressing the puzzle of systematicity for neural networks.  Fodor and Pylyshyn's challenge is still alive and kicking, and foundational philosophical debates from the nineteen-eighties about the scopes of symbolic and connectionist models are as relevant today as when they were first articulated.

\section{Meta-learning for compositionality}
\subsection{Details of the language task}
\begin{table*}[t]
  \centering
  \begin{subtable}[t]{0.3\textwidth}
    \centering
    \begin{tabular}[b]{cc}
      \toprule
      Input        & Output                                   \\\midrule
      dax          & {\color{red} \(\bullet\)}                \\
      fep          & {\color{blue}\(\bullet\)}                \\
      kiki         & {\color{green}\(\bullet\)}               \\
      dax blicket  & {\color{red} \(\bullet\bullet\bullet\)}  \\
      fep blicket  & {\color{blue} \(\bullet\bullet\bullet\)} \\
      kiki blicket & ?                                        \\
      \bottomrule
    \end{tabular}
    \caption{Examples of strings and their translation in a specific language}
    \label{simple-list}
  \end{subtable}
  \hspace{1em}
  \begin{subtable}[t]{0.65\textwidth}
    \centering
    \begin{tabular}[b]{cc}
      \toprule
      Input & kiki blicket \mybar
              dax $\rightarrow$ {\color{red} \(\bullet\)}\mybar
              fep $\rightarrow$ {\color{blue}\(\bullet\)}\mybar
              kiki $\rightarrow$ {\color{green}\(\bullet\)}\mybar\\
            & dax blicket $\rightarrow$ {\color{red}
              \(\bullet\bullet\bullet\)}\mybar
              fep blicket $\rightarrow$ {\color{blue}
              \(\bullet\bullet\bullet\)}\\
      Expected output & {\color{green}\(\bullet\bullet\bullet\)}\\
      \bottomrule
    \end{tabular}
    \caption{The actual input/output pair seen by the neural network. The first string passed is the query.}
    \label{collapsed-list}
  \end{subtable}
  \caption{A possible training-set item with query, matching input strings in an artificial language to their interpretations as strings of colored discs, leaving the last item (the query) blank.  In (a) we show a list of the input-output mappings for this item as one would most naturally use in ``standard'' meta-learning: a list of individually processed input-output mappings, with a test item.  In (b) we show how this list of mappings is collapsed into a single input-output pair in MLC: the multiple items in (a) are concatenated into a single string with a separator symbol ($\mybar$), and this entire string is the input-side of a single input-output mapping, where the output is the expected response for a substring left unconnected to an output in the input (\texttt{kiki blicket} in this case).}
  \label{tab:a-grammar}
\end{table*}

To demonstrate their solution to the systematicity challenge, \textcite{lake-human-like-2023} investigate a specific sequence to sequence mapping task, where strings in simple artificial languages are mapped to their semantic interpretations.
They train a standard transformer model \parencite{Vaswani:2017} on this specific task using their Meta-Learning for Compositionality protocol (MLC), which we will examine in detail in section~\ref{mlc-technique}.

Each member of the family of artificial languages that they use contains \emph{content words} denoting colored discs and \emph{function words} denoting transformations of adjacent meanings.\footnote{We use the term ``function word'' here in the sense familiar from mathematics, rather than linguistics: these are words whose effect is to take one or more arguments and return some transformation of the interpretations of those arguments.}
For example, one such language might have a content word \texttt{fep} which means {\color{blue}$\bullet$} and a function word \texttt{blicket} which means ``repeat the meaning just before three times,'' so that \texttt{fep blicket} would mean {\color{blue}$\bullet \bullet \bullet$}.
Table~\ref{tab:a-grammar} shows a sample meta-learning item in detail.

All training grammars consisted of precisely four content words and precisely three function words, for example words which repeat the preceding substring $n$ times or flip the order of two substrings.
Each function word took either one or two arguments, always adjacent substrings.
These arguments were always fixed to either be substrings of arbitrary complexity themselves or ``atomic'' content words denoting a single colored disc.
Function words were only allowed to ``return'' sequences of at least 2 discs and at most 8 discs.
So ``repeat the meaning of the word before 5 times'' was part of the training set, but ``repeat the meaning of the word before 9 times'' was not.

\textcite{lake-human-like-2023} also conducted a behavioral experiment on human participants.
They asked subjects to perform the same task as the models at the evaluation stage: inferring an output string from a set of input-output pairs plus a query.
Human participants did this with no analog of the MLC protocol, that is no meta-learning of any kind, the natural hypothesis being that humans already think systematically.
In the main study Lake and Baroni report, all evaluations were conducted on a small subset of grammars, dubbed the ``gold grammars,'' which have three specific function-word meanings, and differ only with regard to which specific meanings from this common pool are assigned to which particular labels in a grammar.
We describe this as a \emph{grammar blueprint} as it defines the functional rules but not the specific color meanings or the specific words of a grammar.
The three rules of the gold grammar are ``after'' (flip the meanings of the two substrings on either side, i.e.\ $x_1\ f\ x_2 \to x_2 x_1$), ``thrice'' (repeat the meaning of the preceding content word three times, i.e.\ $u_1\ f \to u_1 u_1 u_1$) and ``surround'' (surround the meaning of the content word to the right with two instances of the meaning of the content word to the left, i.e.\ $u_1\ f\ u_2 \to u_1 u_2 u_1$).

Crucially, when learning a grammar, one content word occurs only in isolation with its meaning, never occurring in complex strings with any of the function words.
Then, the query string uses one or more function words with this held-out content word, in a novel composition not seen in the evaluation set.
If the model behaves systematically, it ought to correctly generalize the use of the function words on a query of novel complexity including the held-out content word.

\subsection{Meta-learning for compositionality: the technique}
\label{mlc-technique}
\textcite{lake-human-like-2023} present a novel technique they name Meta-Learning for Compositionality (MLC), which uses a standard neural network architecture to \emph{meta-learn} over a family of tasks.
We will focus here exclusively on sequence-to-sequence tasks representing language learning, since this is the example Lake and Baroni use as well, but our discussion here, like Lake and Baroni's, extends to other kinds of tasks.
We describe MLC partly in contrast with Model-Agnostic Meta-learning (MAML; \citeauthor{finn_model-agnostic_2017}, \citeyear{finn_model-agnostic_2017}), to provide a more familiar anchor to the reader.

Common meta-learning techniques such as MAML use second-order gradients to adapt the initial parameters of a neural network by doing gradient descent on a task and then updating the initial parameters of a model.
The model is trained directly on many different sequence-to-sequence tasks, optimizing its parameters through many such cycles.
After meta-learning in this fashion, the model is trained further on a specific task using gradient descent, and then tested.\footnote{There is some work in cognitive science using MAML as a way of instilling in a neural network biases meant to help with generalization tasks.
  For example, \textcite{McCoy:2023} use MAML to meta-train a neural network on formal languages, and demonstrate that the model, when later trained on English, shows some benefits to depth generalization.
  \textcite{Binz:2023} review this and other successes of MAML in cognitive science as a means of instilling (chiefly Bayesian) biases in neural networks, while \textcite{Goodale:2025} challenge the interpretation of these results, arguing that MAML can be effective at instilling certain computational mechanisms (e.g.\ counters), rather than \emph{specific} prior distributions.}

\begin{figure*}[t]
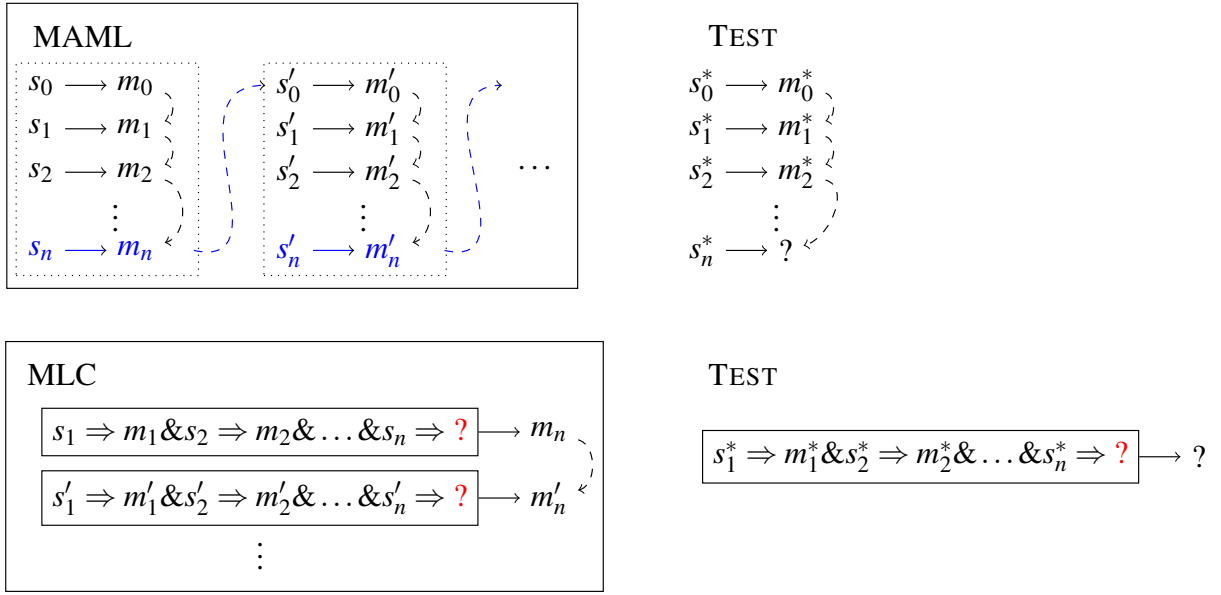

  \centering
  \includestandalone[width=\textwidth]{schematic}
  \caption{Schematic comparison of MAML and MLC\@.  In MAML there are multiple cycles at meta-learning, within each of which there is incremental learning from multiple input-output mappings of strings ($s_i$) to meanings ($m_i$).  At the end of each cycle the product of learning in that cycle is used to adjust the initial parameters for the next cycle.  At the test phase, a new series of strings and meanings ($s^*_i, m^*_i$) is learned incrementally, and the model's prediction for a novel string ($s^*_n$) is checked.  In MLC there is only one cycle of meta-learning, where each input is a long string encoding multiple individual examples of string\hyp{}to\hyp{}meaning mappings, with a query (red question mark).  The outputs are the correct meaning values for the missing queried element.  At testing, there is no learning by gradient descent, a new long string encoding multiple individual examples is provided as input, and its output is checked.}
  \label{fig:meta-learnings}
\end{figure*}

MLC is quite different from MAML as just summarized, see Figure~\ref{fig:meta-learnings} for a direct schematic comparison.
In MLC, there are no analogous multiple cycles of meta-learning.
Instead, MLC uses a single sequence-to-sequence task at the meta-learning phase, where each input sequence effectively corresponds to a cycle of MAML\@.
That is, what was, in MAML (or other meta-learning paradigms), a collection of sequence\hyp{}to\hyp{}sequence mappings from which the network learned in the standard incremental fashion, MLC collapses into one large string (see Table~\ref{tab:a-grammar} for a concrete example of such a collapsed string).
Moreover, at the testing phase, MLC directly presents the network with an input sequence which also collapses multiple input-output mappings into one large string.

Consequently, MLC is much more limited than typical meta-learning approaches, both at the meta-learning phase and at the test phase.
Meta-learning typically involves optimizing both the specific task and the learning of that task.
In MAML, this is operationalized by updating the initial parameters of the model using a held-out test set for each task.
MLC, however, makes use of only one level of learning.
Moreover, the one level of learning found in MLC must handle extremely complex sequences, as it were, \emph{holistically}.
This is a consequence of the strategy of taking a list of input-output mappings and collapsing it into a string to be taken as input.

When learning a new task after the meta-learning training, MAML typically gets to benefit from yet another cycle of learning with gradient descent.
The network is tested by showing it multiple examples of a new kind of input-output sequence, it gets to learn from these examples, and then it must predict the output for a given input.
MLC is much more limited with novel tasks: instead of getting to learn from a few examples, here too it sees these examples collapsed into one long string, taken as an input.
This means that, after meta-learning, the network never again adapts its parameters during testing.
In other words, after meta-learning, the network is essentially tested as in a one-shot learning experiment.

In sum, Meta-Learning for Compositionality might in fact be more appositely classified as a kind of \emph{pretraining}: what conceptually corresponds to multiple-language learning tasks, each with a number of examples, gets collapsed into one large learning task on which the network is pretrained, to then be tested directly, with no additional training.

\subsection{MLC in out-of-distribution tasks}

It is not entirely clear how one would use MLC on a novel task which had different symbols than those encountered in training, as there is no gradient descent or other method at testing to integrate novel symbols.
This means that MLC may be limited strictly to the \emph{specific} task that the meta-training was conducted on. 
This greatly hinders MLC's potential for applications, in AI as in cognitive science, since human intelligence produces systematic behavior even in entirely novel tasks, as \citeauthor{lake-human-like-2023}'s (\citeyear{lake-human-like-2023}) own behavioral experiment attests.

Conversely, MAML can be applied to different tasks, even ones with different kinds of inputs.
While the weights of an MLC model aren't updated at test time (see the right hand side of Figure~\ref{fig:meta-learnings}), with MAML it is entirely possible to add or change parts of the weights and still have them updated in gradient descent at test time.
The simplest example of this is \emph{tokenization}. 
\citeauthor{lake-human-like-2023}'s model has a small vocabulary of nonce-words and colors, no other words or symbols can be added to the model, as there is no representation of them at test time, nor crucially a way to learn to represent new symbols.
By contrast, it is very straightforward to modify the vocabulary of a MAML model to add new symbols, by expanding the embedding matrix and learning the new symbols via gradient descent at test time.
Although different levels of success will be observed depending for example on the richness of the new examples, the protocol has a perfectly clear avenue to leverage gradient descent above and beyond the meta-learning phase.
For example, \textcite{McCoy:2023} use MAML to train an LSTM on various formal languages with a small, fixed vocabulary.
They then train this meta-trained LSTM on an English corpus, meaning the LSTM must generalize at test time on the basis of completely new tokens.
Since MLC produces a transformer with frozen weights and vocabularies, it could not perform this vital task.

% While it is not remarked upon by \textcite{Fodor:1988}, nonce words are a crucial example of systematicity in natural language.
% Even if you have never seen a given noun before, there is no syntactic position for nouns where you cannot use it.
% For example, if I point to an animal and call it a ``polyphaunt'', you could very easily use it in any kind of sentence despite not knowing the word beforehand (e.g.\ ``I saw a polyphaunt'', ``Polyphaunts are a kind of animal'', ``I am not sure how many polyphaunts there are'', etc.)
% Conversely, with MLC, the model is simply incapable of using a nonce word which is not included during meta-training.
% MAML and other meta-learning techniques can acquire new words that are unused during meta-training (although they have not been shown to do so in a systematic manner).

One possibility, briefly considered by the authors in the discussion section, is that young children engage in many different tasks requiring systematicity, and therefore may \emph{learn} systematic behavior by getting an analog of meta-training for compositionality with these different tasks performed very early in their lives.
While the authors do not describe how precisely this could be done with the MLC protocol, the most natural possibility is to administer MLC over an extremely diverse set of tasks.
Large Language Models (LLMs) could be seen as a straightforward case study, for these models are often taken to have learned a diverse set of tasks indirectly while being trained ``merely'' on next-word prediction.\footnote{We hasten to add that we do not necessarily subscribe to this interpretation, and instead remain agnostic as to whether it is meaningful to say, as many do, that state-of-the-art LLMs were trained ``merely'' on next-word prediction yet acquired an ability to in-context learn a host of other tasks like text summation, differential medical diagnosis, psychoanalysis, or just about any task that can be operationalized in natural language.  In many of our moods, we think that there is nothing ``mere'' about next-word prediction when the language at hand is a natural language, the unbounded representational system \emph{par excellence}, and the training set includes just about the totality of recorded human intellectual activity.}
Indeed, the way an MLC-trained model predicts the output given example strings bears a striking resemblance to in-context learning in LLMs.
In-context learning is a phenomenon observed in LLMs where a model can perform an arguably novel task on the basis of a single or a few examples, provided as a text prompt.
For both the MLC-trained network and LLMs, when the model must infer an answer in context, it must do whatever ``learning'' it needs via its fixed weights, without any gradient descent.
However, \citeauthor{lake-human-like-2023}'s (\citeyear{lake-human-like-2023}) own supplementary materials pour cold water over this intriguing possibility.
Despite reams and reams of pretraining arguably on a broad range of compositional tasks, GPT-4 is only able to achieve $58\%$ accuracy on the authors' task.

% It seems unlikely that MLC should be able to generalize to other systematic tasks beyond what is present in its meta-training dataset.
% Curiously, recent work has analyzed ICL using neurosymbolic methods where an untrained neural network is designed to implement a specific symbolic program (SMOLENSKY).
% This approach would be entirely consonant with \citeauthor{Fodor:1988}'s (\citeyear{Fodor:1988}) description of neural networks as a potential implementations of symbolic theories.

\section{Failures of systematic generalization}

Beyond the idiosyncrasies of the MLC approach we have described, there remains an open question as to whether MLC succeeds on its own narrowly behavioral terms.
That is, under the constraints of the very particular task and evaluation set up by \textcite{lake-human-like-2023}, does the model behave systematically?
To answer this question, we carefully evaluated their ``gold grammar'' as well as similar language blueprints with small changes to the rules.
We found failures of systematicity that show that \citeauthor{lake-human-like-2023}'s (\citeyear{lake-human-like-2023}) model fails even on their own behavior-centered criteria.

We classify these errors into three levels, in increasing (qualitative) level of abstraction.
Level one failures are \emph{within-grammar} failures, where the model succeeds at correctly interpreting a string but fails for a minor modification of it.
Level two considerations look \emph{across instantiations of a grammar blueprint} at correlations between network performance and properties of grammars which oughtn't to matter, specifically what labels (words) are mapped to what colors (meanings).
At level three we look \emph{across grammar blueprints} at the learning profiles for different rules.
Our observations here concern the \texttt{algebraic} model, which had the best generalization performance, but these problems are present with the other variants too.

To discuss the gold grammar and our variants of it without too jarring a profusion of symbolic notation, in what follows we use $c_i$ both to represent a specific color meaning \emph{and} the content word for that color.
Symbols like $c_1$, $c_2$, $c_3$ are the colors that were used with function words in training examples, and $c_h$ is the held-out color used to test generalization.
For example, ``$c_{1}$ surround $c_{h}$'' might be instantiated in a particular language as \texttt{dax kiki fep}, where \texttt{dax} would be a word seen during training as the argument of at least one function word, and \texttt{fep} would be a word seen in training only once and by itself, that is without being the argument of any function words.
This string would have the schematic meaning $c_1 c_{h} c_{1}$, as the function word, \texttt{kiki}, means ``surround.''

For each level of failure, we conducted different experiments which found short-comings of the \texttt{net-BIML-algebraic-top.pt} model trained by Lake and Baroni.
We chose to analyse this model as it had the highest behavioral performance and was reported as being ``perfectly systematic.''
Unless otherwise indicated, all hyperparameters, training and model details were identical to those of Lake and Baroni's paper.
Furthermore, we illustrate that changing the distribution of meta-training data even slightly can lead to blatantly non-systematic models.
All code for our experiments are available on GitHub at \href{https://github.com/MichaelGoodale/mlc-reply}{MichaelGoodale/mlc-reply}.

\subsection{Level one failures}
Lake and Baroni report several successfully generalized strings, but we found the model fails for strings of comparable complexity.
We quickly found the schema ``$c_1$ surround $c_h$ after $c_h$ thrice,'' which is incorrectly predicted in many grammars (see Table~\ref{tab:level2-word-sensitive-rules} and the supplementary materials).
Importantly, this schema is a very minor variation of one tested in the article and successfully learned within the same grammars that fail our variant: ``$c_h$ surround $c_h$ after $c_h$ thrice.''
The two schemata are different special cases of ``$x$ surround $y$ after $z$ thrice,'' showing that performance differs dramatically between equally valid  combinations of $x$, $y$, $z$ ($c_h$, $c_h$, $c_h$ vs.~$c_1$, $c_h$, $c_h$).
This is akin to a human understanding the English sentence  ``Ann introduced Bill to Claire'' while being stumped by ``Dan introduced Claire to Anne,'' betraying a deep failure of systematicity in its most paradigmatic manifestation.

\begin{figure}
  \includegraphics[width=\columnwidth]{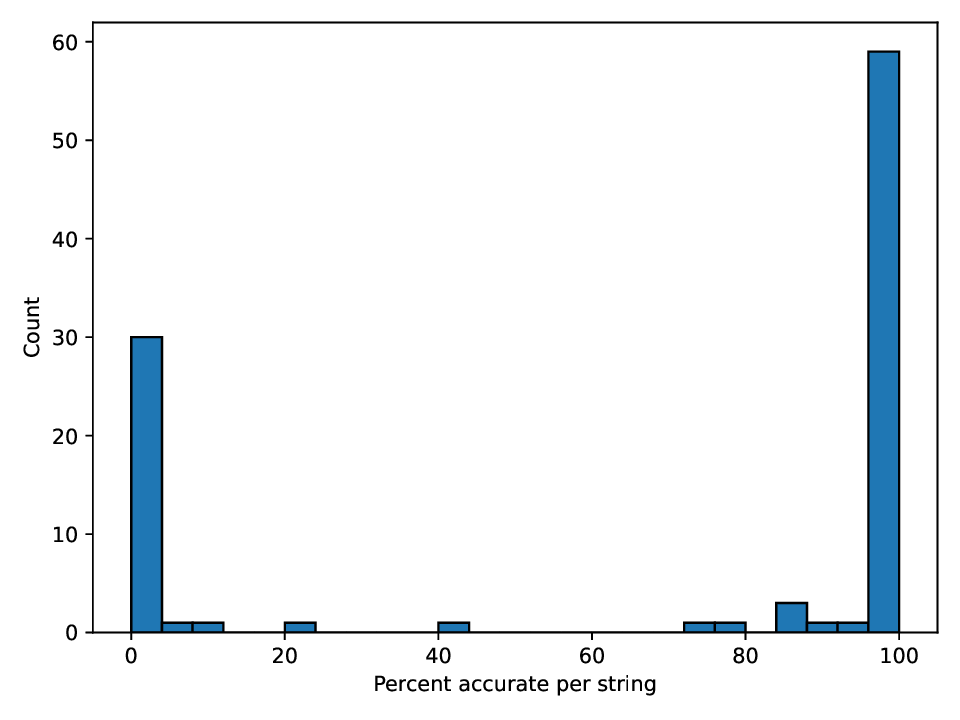}
  \caption{Distribution of performance accuracy accross all label-color pairs for 100 random strings in the gold grammar. \label{fig:hist}}
\end{figure}

To find these strings, we used Lake and Baroni's own code to generate novel test examples which perfectly mimicked their evaluation on the gold grammar while randomly changing the complex query string.
That is, we use the exact same example strings (modulo colors/labels) as Lake and Baroni, but evaluate on a different novel string.
The model was then run on all possible label-color pairs for the same example and query strings of the gold grammar.
We then looked at the model's performance on different strings in the gold grammar generated using seeds $1$ to $100$.
Many strings were inaccurately predicted, out of all seeds from 1 to 100, only $51$ had perfect performance across all label-color pairs, $24$ had less than perfect performance and $25$ had 0.0\% accuracy (see the distribution in Figure~\ref{fig:hist}). 
We chose to look at ``$c_1$ surround $c_h$ after $c_h$ thrice'' in detail, because of its close similarity to Lake and Baroni's own evaluation string.

\begin{table*}[t]
  \centering
  \begin{tabular}{lcc}
    \toprule
    & \multicolumn{2}{c}{Count across label-color pairs}\\
    Output & Static order & Shuffled order\\
    \midrule
    $c_h$ $c_h$ $c_h$ $c_1$ $c_h$ $c_1$ &  1993 & 2785\\
    $c_1$ $c_h$ $c_h$ $c_h$ $c_h$ $c_1$ &  1045 & 934\\
    $c_h$ $c_h$ $c_h$ $c_h$ $c_1$ $c_1$ &  1282 & 598\\
    $c_h$ $c_h$ $c_h$ $c_h$ $c_1$ $c_h$ &  0    & 3 \\
    \bottomrule
  \end{tabular}
  \caption{Four very different outputs for the same schema ``$c_1$ surround $c_{h}$ after $c_h$ thrice,'' with counts across label-color pairings.
    The first is the normatively correct output.
    The second is a possible output if one parses the string as ``$x$ surround ($y$ after $z$ thrice),'' though technically this was unintended with the gold grammar, as ``surround'' examples only ever take two primitives (color-denoting labels) as arguments.
    The third and fourth aren't acceptable outputs under any parse.
    \emph{Static order} vs.~\emph{shuffled order} concerns the order in which the study items were presented to the network.
    With static order of presentation the network's accuracy on this schema is at most 70\%, with shuffled order at most 86\%.
    Brenden Lake (p.c.) suggested that we try shuffling the presentation order of the study items to ensure that this wasn't a idiosyncratic failure of the model for that particular order.
    The different results depending on order of presentation are instructive, as they indicate their own kind of non-systematic fragility.
  }
  \label{tab:level2-word-sensitive-rules}
\end{table*}

\subsection{Level two failures}
The model is sensitive to precisely which labels (words) are mapped to which meanings (rules and colors), with different mappings producing different performances.
Thus, the networks do not conform to the arbitrariness of linguistic signs, whereby no particular benefits or drawbacks are created by American English ``sofa'' vs.~(Old) Canadian English ``chesterfield.''

For example, the same schema discussed above (``$c_1$ surround $c_h$ after $c_h$ thrice'') shows different learning successes \emph{across} grammars, being correctly generalized for some label-color pairs and not for others (Table~\ref{tab:level2-word-sensitive-rules}).
Additionally, we found that the model has different success rates on the schema, depending on precisely which color is held out (Table~\ref{tab:level2-results-by-withheld-color}).

To produce these data, we simply took our complex string from before, and evaluated the model's performance on different label-color pairs, finding odd patterns such as worse performance if yellow is the held-out color.

\begin{table*}
  \centering
  \begin{tabular}{lccccc}
    \toprule
    & \multicolumn{2}{c}{Constant presentation order} & & \multicolumn{2}{c}{Shuffled presentation order}\\
    Held out color & \% correct & SEM & & \% correct  & SEM\\
    \midrule
    Blue            & 56.11      & 1.8506 & & 59.72 & 1.8291\\
    Green           & 62.08      & 1.8094 & & 73.47 & 1.6464\\
    Pink            & 38.47      & 1.8144 & & 67.36 & 1.7487\\
    Purple          & 50.28      & 1.8647 & & 69.03 & 1.7244\\
    Red             & 53.75      & 1.8594 & & 75.83 & 1.5965\\
    Yellow          & 16.11      & 1.3710 & & 41.39 & 1.8368\\
    \bottomrule
  \end{tabular}
  \caption{Accuracy on held-out-label-color combinations for the string ``$c_1$ surround $c_h$ after $c_h$ thrice.''}\label{tab:level2-results-by-withheld-color}
\end{table*}

\subsection{Level three failures}
\begin{figure*}
  \centering
  \includegraphics[width=\textwidth]{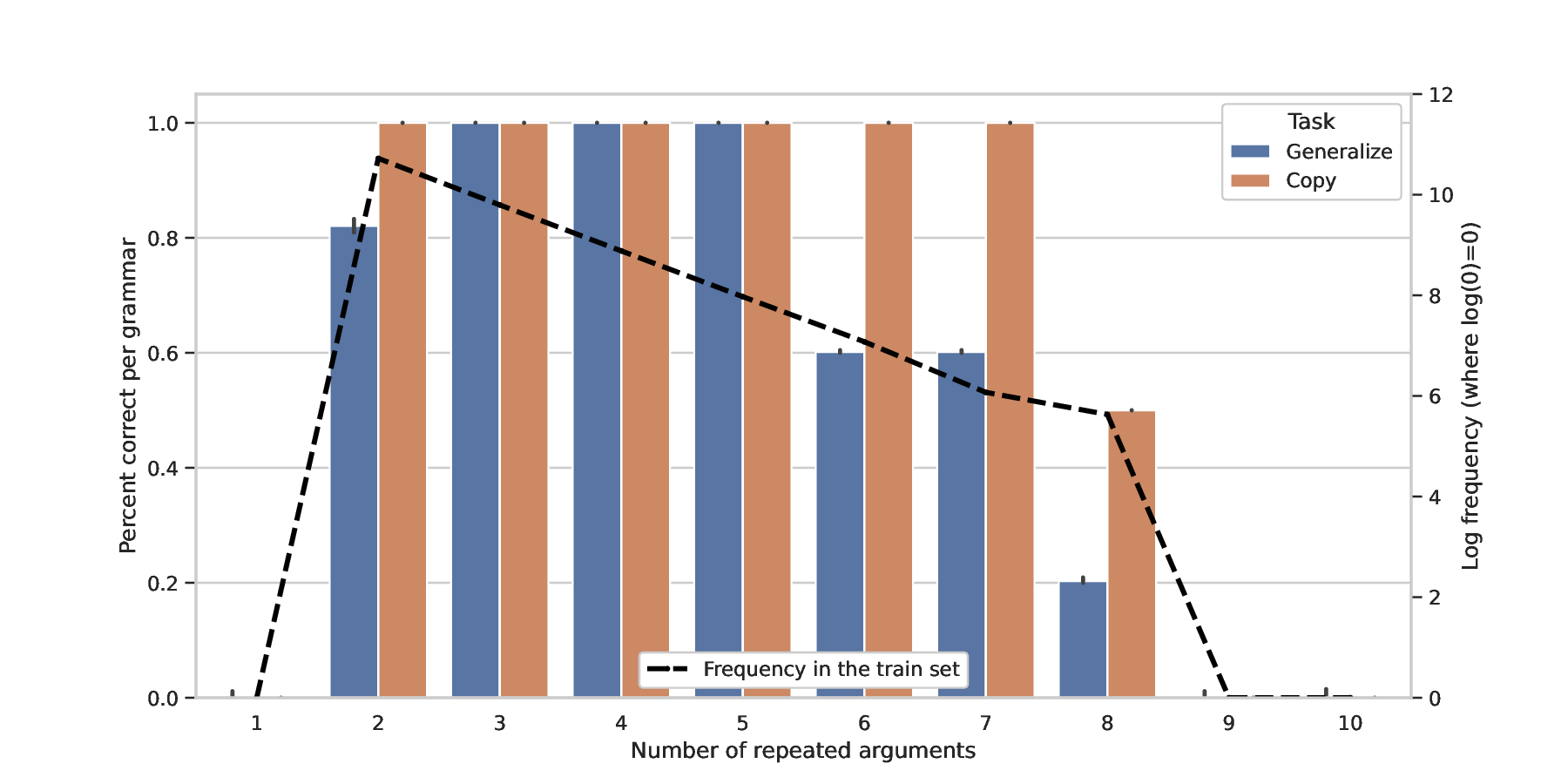}
  \caption{Learning functions that take an argument $x$ and return $n$ repetitions of $x$.
    \emph{Generalize} strings are the original test sequences where we replace thrice by $n$.
    \emph{Copy} shows the model's ability to repeat sequences that are provided as study examples (i.e.~lookup).
    The figure shows confidence intervals but the distribution has so little variation that they are hard to see.}
  \label{fig:repeater-function}
\end{figure*}

We found that the network is overly sensitive to what it saw at the meta-learning phase.
The meta-learning protocol only exposed the networks to rules whose outputs had lengths between 2 and 8 words, and we found that they could not learn a rule that was a trivial extension, violating this 2--8 constraint.
Specifically, while the model can learn a function that takes one argument and repeats it $n$ times for $2 \le n \le 8$, it couldn't learn an analogous rule that repeats its argument, say, 9 times.
This cannot be due to traditional constraints on production such as limited memory, as the model was unable to learn to repeat its argument \emph{once}, showing it failed to learn even the identity function.
This shows that the model isn't a \emph{systematic learner}, instead, it is entirely beholden to what precisely appeared in the meta-learning phase.
Additionally, we found that performance within the 2--8 length bounds drops as the output gets longer, mirroring the frequency with which functions  were encountered at meta-learning (see Figure~\ref{fig:repeater-function})

Perhaps most damningly, the model fails to generalize to new lengths even on a simple copying task.
Copying requires providing the output of an input string which is in the prompt as an input-output pair. 
In other words, it requires nothing more than the ability to copy a string.
Yet, the model is still incapable of doing it when provided any rule outside the 2--8 length range.

To show this, we generated a new dataset, varying the gold grammar in a systematic way as before.
However, instead of changing the complex query string, we systematically manipulated the meaning of the rule ``thrice,'' changing it from copying its input three times, and generalizing it to different $n$ (e.g.\ repeat once, twice, etc).
As shown in Figure~\ref{fig:repeater-function}, the model is unable to learn rules where $n$ is not between 2 and 8, and shows a gradual drop off as the number of repetitions increases within 2 to 8. 

\begin{figure}
  \includegraphics[width=\columnwidth]{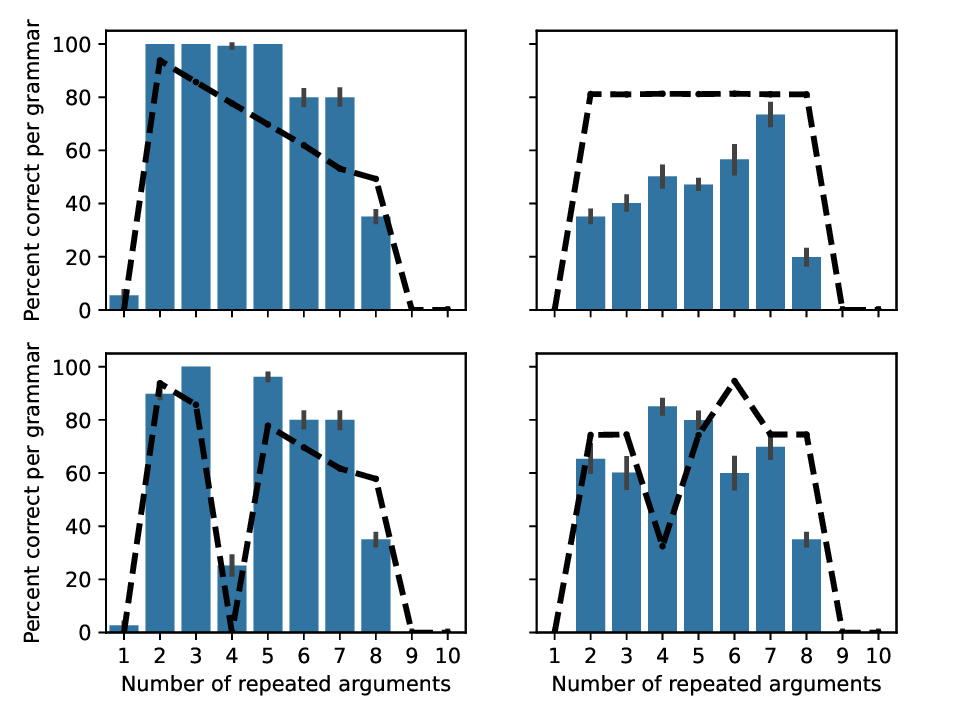}

  \caption{
    An expansion of Figure~\ref{fig:repeater-function}  looking only at  the performance on \textit{Generalize} tasks, depending on what distribution of rules the MLC model is meta-trained on.
    The frequency of each rule for different MLC models is shown with the dashed line.
    Top left is a MLC model trained on a novel dataset that uses the same distribution as used by Lake and Baroni, replicating the results in Figure~\ref{fig:repeater-function}.
    \label{fig:different}
  }
\end{figure}

At a reviewer's helpful suggestion, we trained several new models while systematically changing the distribution of the lengths of the output of rules.
Lake and Baroni's MLC dataset randomly generated rules by permuting (with repetitions) the arguments of the function. 
The number of times the arguments were repeated was determined using a geometric distribution clamped between 2 and 8.
We manipulated the distribution of lengths to produce models with different biases towards different kinds of rules as shown in Figure~\ref{fig:different}. 
We found that a uniform distribution over lengths (2--8) led to a model which struggled to generalize altogether.
If we excluded rules of length 4, the model was unable to reliably generalize for those rules, showing an inability to interpolate between rules.
These results show the MLC model's deep difficulty with generalization to novel kinds of rules and a deep sensitivity to the particularities of the meta-training dataset.

\section{Discussion and conclusions}
We found three different levels of failures of systematic generalization in Lake and Baroni's algebraic network.
We conclude that, while the network did show some measure of success at later learning on the basis of extremely small datasets, these successes cannot be described as instances of systematic generalization, so that the systematicity challenge is in fact \emph{not} addressed by the MLC protocol in its current shape, even if we (erroneously) take the systematicity challenge to be exclusively about generalization in behavior.

One counter to our conclusion, consonant with the methodology of the article which included behavioral data from humans, is to propose that neural networks need only be as systematic as humans are \emph{in performance}.
There are all manner of reasons why a particular human in a particular circumstance might behave differently on the basis of the two stimuli ``Ann introduced Bill to Claire'' and ``Claire introduced Dan to Ed.''
Having had an unpleasant experience with someone named Bill might do the trick.
Cognitive science recognizes that a faculty may be systematic, while particular circumstances of use of a faculty might not show it.

Thus, one could in principle accept the classical points about systematicity, despite performance complications, and somehow investigate the \emph{competence} of MLC networks, to show that \emph{it} is systematic in the required sense \parencite{Firestone:2020}.
At points, the authors imply that what they call the algebraic model is a model of competence, while the models supplemented with heuristics and human behavioral data would be models of performance.
Focusing on the question of competence and accordingly focusing on the algebraic model does not help in this connection, for our discoveries here concern precisely the algebraic model, and show that it is not systematic.

But taking this work as being strictly about performance introduces its own problems.
For example, we are told that ``the \emph{top-performing} MLC [\ldots]\ was jointly optimized on both the few-shot learning task and the open-ended human responses'' (Lake and Baroni, 2023, p.~4, our emphasis) indicating that the hallmark of good performance for the authors is reproduction of human behavioral patterns, regardless of normative generalization or strict systematicity.
But if this species of pure imitation game \parencite{Turing:1950} is the real goal here, then the rhetoric elsewhere in the article becomes difficult to understand.
Everywhere else, the authors engage directly with Fodor and Pylyshyn, yet it is impossible to interpret the systematicity challenge as a challenge about \emph{imitation}.
The whole notion of systematicity is precisely about properties of the system that \emph{generates} behavior, not the imitation of said behavior, a point acutely clear to anyone who participated in the cognitive revolution as Fodor and Pylyshyn did.
Moreover, it is worth pointing out that the meta-training of this model is done on the very task and class of grammars that occur at testing, and then optimized on human behavioral data from the very same task and class of grammars.\footnote{The gold grammars were of course held out from the meta-training protocol, but note that each of the content words, function words, and sentences that occur in the gold grammar at testing were all part of the distribution at training.  That is, each of the subparts of the gold grammar is entirely within the training distribution.}
It is therefore not exactly surprising that this model should display behavior in this task and class of grammars that imitates the human data quite closely.

Such a restriction to human-like \emph{behavioral} systematicity altogether misses the \emph{stated} objective (and the stated outcome) of the work: to ``successfully address Fodor and Pylyshyn’s challenge'' (Lake and Baroni, 2023, abstract).
Even if Lake and Baroni's method had achieved generalization powers entirely identical to those of their human participants in the experimental conditions they define, it would remain deeply misleading to characterize such ``success'' as addressing the systematicity challenge.
We think that this matters greatly, because we see real value in the systematicity challenge, as even some of the most influential scholars on the opposing side recognized at the time \parencite{Smolensky:1987}.
It is of course entirely valid to choose \emph{not} to address the challenge of systematicity, and focus instead on purely behavioral imitation and behaviorally-accurate generalization.
But we believe that it is very problematic to claim to address the challenge, while in fact misrepresenting it.
This has the effect, compounded when publication occurs in our profession's flagship journal, of misleading the readership into thinking that an important challenge has been met and resolved, potentially stymieing progress on that challenge.

Moreover, our experiments here demonstrate that Lake and Baroni pronounced their mission accomplished too early, even granting their inaccurate portrayal of the systematicity challenge as a purely behavioral one.
Their human-subjects experiment tested participants on the very same highly specific test case, with only a handful of complex test strings, so that the many novel failures of behavioral systematicity which we uncovered in the MLC algebraic model were not tested with humans.
Perhaps humans will behave just like Lake and Baroni's models in our new MLC experiments, though we strongly suspect they will not.
For instance, we submit that any cognitive scientist would be befuddled to learn that humans display the performance patterns in our Figure~\ref{fig:different}.
In particular, if examples of the repeating function (``thrice'' in the gold grammar) at training omitted the $n = 4$ case (bottom-left panel of Figure~\ref{fig:different}), we would fully expect humans to interpolate the possibility of this held-out example with no difficulty, unlike the MLC-trained models.
Be that as it may, the onus must rest squarely with the proponents of MLC to investigate the behavioral generalization abilities of their models extensively enough before pronouncing a decades-old challenge resolved.

What remains clear, however much weight one is inclined to put on Fodor and Pylyshyn's argument from systematicity, is that their challenge still stands, and that meta-learning (for compositionality or otherwise), while interesting and potentially extremely useful, has yet to fulfill its promise to bridge this particularly important gap between symbolic and connectionist approaches.
Current neural networks with generic architectures, that is models without hard and strong prior constraints tailored to guarantee emulation of the core properties of symbolic systems, neither produce systematic behavior nor explain how systematicity can be a fundamental property of minds.

\section*{Acknowledgments}
For extremely helpful discussion, we thank Yair Lakretz, Jon Rawski, Natasha Korotkova, and the audiences of the LANG-REASON group at Ecole Normale Supérieure, and the 25th Szklarska Poręba Workshop on the Roots of Pragmasemantics, as well as the editor and reviewers of TACL.
We also thank Brenden Lake for a brief but very helpful email exchange at an early stage of this work. 

\section*{Author contribution statements}
Both authors were involved in the study's conception and design.
Programming and running experiments was handled by MG\@.
Both authors contributed to the writing and editing of the article.

\section*{Code availability}
All relevant code for these experiments, allowing for full reproduction of our results, can be found at \url{https://github.com/MichaelGoodale/mlc-reply}.

\section*{Data availability}
All relevant data is included in the repository of our code or can be re-created by running our code.

\section*{Funding acknowledgment}
The work reported here was funded in part by Agence Nationale de la Recherche grant ANR-19-P3IA-0001 (PRAIRIE 3IA Institute, PI: Mascarenhas) and by a grant from Ecole Doctorale Frontières de l'Innovation en Recherche et Education---Programme Bettencourt (PhD funding for Goodale).

\bibliography{references}
\bibliographystyle{acl_natbib}

\end{document}

% LocalWords:  MAML ICL GPT blicket dax fep kiki

%% file: schematic.tex
\begin{tikzpicture}

  \begin{scope}
    \foreach \x in {0,...,2}
      {\node (A\x) at (0, -\x/1.9) {$s_\x$};
      \node (B\x) [right=0.5 of A\x]    {$m_\x$};
      \draw[->,] (A\x) -- (B\x);}

      \node at (0.9, -1.5) {\vdots};
      \begin{scope}[text=blue,draw=blue]
        \node (A3) at (0, -2) {$s_n$};
        \node (B3) [right=0.5 of A3]    {$m_n$};
        \draw[->] (A3) -- (B3); 
      \end{scope}

      \foreach \x/\y in {0/1, 1/2, 2/3}
      {
        \draw[->, dashed] (B\x) to[out=-15, in=15] (B\y);
      }
    \node[draw, dotted, fit={($(A0)-(5pt, -4pt)$)  ($(B3) +(18pt, -5pt)$)}] {};
  \end{scope}

    \begin{scope}[shift={(3, 0)}]
    \foreach \x in {0,...,2}
      {\node (C\x) at (0, -\x/1.9) {$s_\x'$};
      \node (D\x) [right=0.5 of C\x]    {$m_\x'$};
      \draw[->] (C\x) -- (D\x);}

      \node at (0.9, -1.5) {\vdots};
      \begin{scope}[text=blue,draw=blue]
        \node (C3) at (0, -2) {$s_n'$};
        \node (D3) [right=0.5 of C3]    {$m_n'$};
        \draw[->] (C3) -- (D3); 
      \end{scope}

      \foreach \x/\y in {0/1, 1/2, 2/3}
      {
        \draw[->, dashed] (D\x) to[out=-15, in=15] (D\y);
      }
    \node[draw, dotted, fit={($(C0)-(5pt, -4pt)$)  ($(D3) +(18pt, -5pt)$)}] {};
    \draw [->, draw=blue, dashed] ($(B3.east) + (10pt, 0)$) .. controls (0, -2.25) and (-1.5, 0) .. (C0);
    \begin{scope}[shift={(3,0)}]
      \node (fin) at ($(C0.west)+(3, 0)$) {};
      \draw [->, draw=blue,dashed] ($(D3.east) + (11pt, 0)$) .. controls (0, -2.25) and (-1.5, 0) .. (fin);
      \node (cont) at (0, -1) {\ldots};
    \end{scope}
  \end{scope}

  \node (mamlLabel) at (0.5, 0.6) {MAML};
  \node[draw, fit={(A0) (D3) (cont) (fin) (mamlLabel)}] (maml) {};

  \node (mamlTest) at (8.5, 0.6) {\textsc{Test}};

  \begin{scope}[shift={(8, 0)}]
    \foreach \x in {0,...,2}
      {\node (A\x) at (0, -\x/1.9) {$s_\x^*$};
      \node (B\x) [right=0.5 of A\x]    {$m_\x^*$};
      \draw[->] (A\x) -- (B\x);}

      \node at (0.9, -1.5) {\vdots};
      \node (A3) at (0, -2) {$s_n^*$};
      \node (B3) [right=0.5of A3]    {?};
      \draw[->] (A3) -- (B3); 

      \foreach \x/\y in {0/1, 1/2, 2/3}
      {
        \draw[->, dashed] (B\x) to[out=-15, in=15] (B\y);
      }
  \end{scope}

  \begin{scope}[shift={(0, -4.5)}] 
    \node (mlcLabel) at (0.25, 1) {MLC};

    \begin{scope}
      \node[anchor=west,draw] (E1) at (0, 0.3) {$s_1\Rightarrow m_1 \& s_2\Rightarrow m_2 \&\ldots\& s_n\Rightarrow {\color{red}?}$};
      \node[anchor=west,draw] (E2) at (0, -0.5) {$s_1'\Rightarrow m_1' \& s_2'\Rightarrow m_2' \&\ldots\& s_n'\Rightarrow {\color{red}?}$};
      
      \node (F1) [right=0.5 of E1] {$m_n$};
      \node (F2) [right=0.5 of E2] {$m_n'$};
      \draw [->] (E1) -- (F1);
      \draw [->] (E2) -- (F2);

      \draw[->,dashed] (F1) to [out=-15, in=15] (F2);

      \node (depth) at (2.65, -1.1) {\vdots};
    \end{scope}

    \node (mclTest) at (8.5, 1) {\textsc{Test}};
    \node[anchor=west,draw] (ETest) at (8, 0) {$s_1^*\Rightarrow m_1^*\&s_2^*\Rightarrow m_2^* \&\ldots\& {s_n^*\Rightarrow {\color{red}?}}$ };
    \node (FTest) [right=0.5 of ETest] {?};
      \draw[->] (ETest) -- (FTest);

    \node[draw, fit={($(E1) + (-5pt, 0)$) (depth) ($(F2.east) + (5pt,0)$) (mlcLabel) }] (mcl) {};
  \end{scope}

\end{tikzpicture}